\documentclass[acmtog,nonacm]{acmart}
\usepackage{lipsum}
\usepackage{booktabs} 

\usepackage{svg}
\usepackage{amsmath}
\usepackage[normalem]{ulem}
\usepackage{xcolor}
\usepackage[utf8]{inputenc}
\usepackage{algorithm}
\usepackage{algpseudocode}

\begin{document}

\title{Adaptive Multi-NeRF: Exploit Efficient Parallelism in Adaptive Multiple Scale Neural Radiance Field Rendering}

\author{Tong Wang}
\affiliation{%
  \institution{Cygames Research, Cygames Inc.}
  \country{Japan}
  }
  
\author{Shuichi Kurabayashi}
\affiliation{%
  \institution{Cygames Research, Cygames Inc.}
  \country{Japan}
  }

  

\renewcommand{\shortauthors}{Wang et al.}

\begin{abstract}
Recent advances in Neural Radiance Fields (NeRF) have demonstrated significant potential for representing 3D scene appearances as implicit neural networks, enabling the synthesis of high-fidelity novel views. However, the lengthy training and rendering process hinders the widespread adoption of this promising technique for real-time rendering applications. To address this issue, we present an effective adaptive multi-NeRF method designed to accelerate the neural rendering process for large scenes with unbalanced workloads due to varying scene complexities.

Our method adaptively subdivides scenes into axis-aligned bounding boxes using a tree hierarchy approach, assigning smaller NeRFs to different-sized subspaces based on the complexity of each scene portion. This ensures the underlying neural representation is specific to a particular part of the scene. We optimize scene subdivision by employing a guidance density grid, which balances representation capability for each Multilayer Perceptron (MLP). Consequently, samples generated by each ray can be sorted and collected for parallel inference, achieving a balanced workload suitable for small MLPs with consistent dimensions for regular and GPU-friendly computations. We aosl demonstrated an efficient NeRF sampling strategy that intrinsically adapts to increase parallelism, utilization, and reduce kernel calls, thereby achieving much higher GPU utilization and accelerating the rendering process. 

\end{abstract}

\keywords{neural rendering, neural radiance fields}
\begin{teaserfigure}
  \centering
  \includegraphics[width=\textwidth]{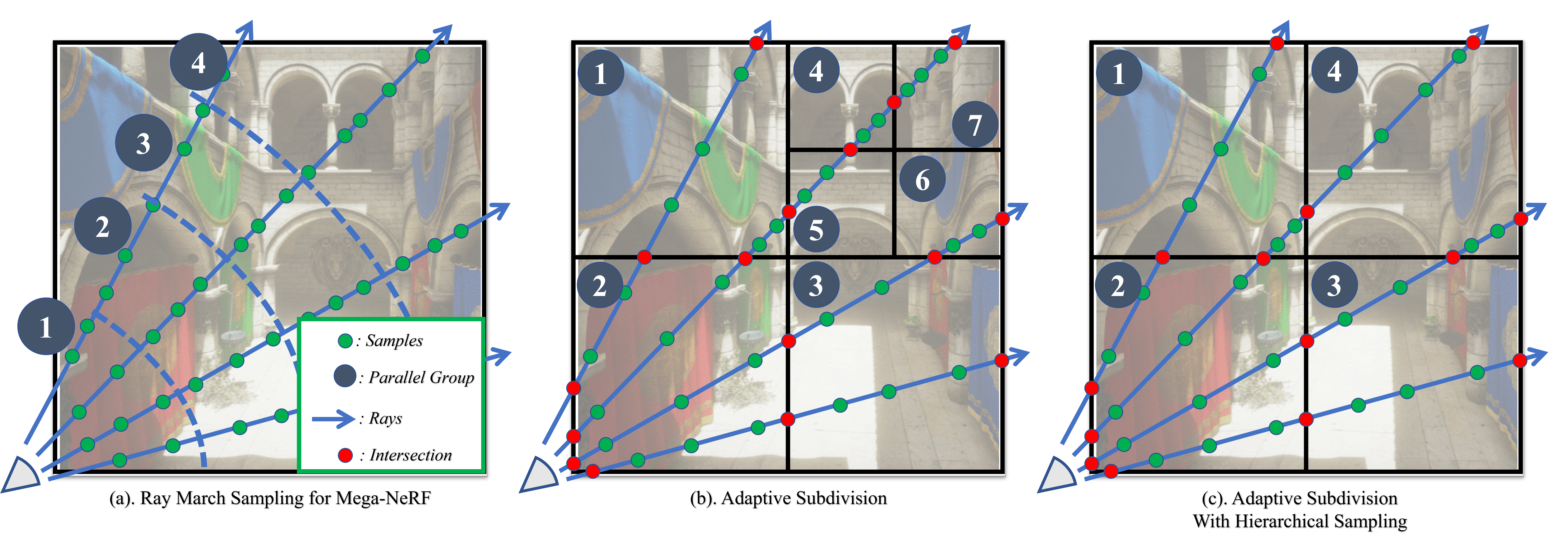}
  \caption{Overview of our adaptive multi-NeRF method: (a) Traditional NeRF sampling and parallelization schemes employ ray marching without scene density awareness, generating samples along the ray, which are then processed by a single, large, and slow-to-infer NeRF MLP. (b) Our adaptive multi-NeRF structure divides the scene into hierarchical axis-aligned bounding boxes (AABBs) using a BSP tree style based on scene density and complexity. Each inner or leaf node is represented by a smaller, faster MLP. Samples are intrinsically adaptive, generated at the intersections of ray sets and the tree structure, allowing samples within one AABB to be batched and inferred with high parallelism. (c) Our adaptive multi-NeRF with hierarchical sampling strategy takes both scene density and camera-to-scene distance into consideration when determining which layer of the BSP tree samples are generated, reducing the number of necessary samples from dense, distant scene parts.}
  \label{fig:teaser_image}
\end{teaserfigure}

\maketitle

\section{Introduction}
 
Neural functions have increasingly become a popular method for parameterizing materials and scene appearances in complex 3D scene synthesis and retrieval. Neural Radiance Fields (NeRF), first proposed in \cite{nerf}, use multilayer perceptrons (MLP) as an implicit volumetric representation of scene appearances, outperforming many other approaches due to their high fidelity representation capability and low storage demands. Despite their high-quality approximation during inference, the computational power required still prevents NeRF from being a practical solution for real-time high-fidelity image synthesis.

A considerable amount of recent literature has focused on high-performance NeRF training and rendering to accelerate the process, with multi-NeRF based methods playing an important role in performance gains. Multi-NeRF methods employ a divide-and-conquer approach to address NeRF's long training and rendering times. While training can be sped up using a multi-GPU cluster, rendering must happen in real-time on a user's device for interactive applications, necessitating instant NeRF synthesis previews.

Inspired by KiloNeRF \cite{kilo}, we also divide the scene into axis-aligned grids and assign a different MLP to each grid for representing appearances within the space. This enables faster computation times during inference compared to using a deep and wide MLP for each sample along each testing ray. However, unlike KiloNeRF, we use the same-sized small network for each grid but allow each grid to represent a different-sized space. This effectively captures the complexity of various scene parts, with some containing more complex features while others have simpler materials and geometry.

The main contributions of this work include the following:

1. The use of adaptive multiple NeRF allows for more efficient representation of the appearance of a scene. By dividing the scene into multiple axis-aligned bounding-boxes and assigning a small MLP to each grid, the irregularity of the scene can be effectively exploited, allowing for a smaller number of NeRFs to be used while still accurately representing the appearance of the scene. This can result in faster training and rendering times and improved quality compared to other multi-NeRF methods. We find potential for adaptive NeRFs to synthesize complex scenes with high fidelity and guaranteed realism, pushing the performance of NeRF rendering one step closer to real-time applications.

2. An adaptive hierarchical sampling and parallel scheduling scheme for effective rendering using our adaptive multi-NeRF approach. During rendering, rays intersect the BSP tree that partitions the scene, and samples are generated, inferred, and composited to form the color of each ray and its corresponding pixel. We adopt an interval-based sampling approach, filling samples within intersecting intervals uniformly, and use a parallel scheduling method to sort and composite sample colors efficiently on GPU for inference. The adaptive nature of our method ensures that regions with similar complexity receive a comparable number of samples, regardless of their spatial dimensions. This adaptivity is demonstrated in the equal distribution of samples in both large and small nodes, as they essentially represent scene parts with similar complexities.

\section{Related Work}

For decades, researchers have been seeking an efficient way to generate photo-realistic images from data representations of concrete scenes. Approaches can be classified based on different representations being explored to generate a novel view for complex material appearance in a free-view angle framework. Traditional computer graphics have exploited various aspects of mesh-based approaches, while image-based methods have seen a recent surge in popularity due to improvements in the study of neural rendering \cite{lombardi_deep_2018,lombardi_mixture_2021,tewari_advances_2022,trevithick_grf_2021,chan_pi-gan_2021,zhang_nerfactor_2021,boss_nerd_2021,neff_donerf_2021}. These advances seek an efficient 3D-structure-aware scene representation that encodes both geometry and appearance, which can be much easier to synthesize than geometry and material modeling from the bottom up.

Novel view synthesis from images has recently seen significant advancements thanks to the use of neural volumetric representations. These new approaches have allowed more accurate and realistic scene reconstructions and have opened up new possibilities for applications in computer graphics and vision. SRNs \cite{SRN} represent scenes as continuous functions that map world coordinates to a feature representation of local scene properties. They can be trained end-to-end from only 2D images and camera poses, without access to depth or shape. This paves the way towards a more precise and high-performance category of using neural networks and a volumetric rendering method to represent and retrieve rendering of novel views. The neural radiance field (NeRF) proposed by \cite{nerf} and its related work have become a fundamental context for most of the following work we are going to discuss.

As presented by \cite{nerf}, NeRF uses a continuous fully-connected deep neural network to represent scenes. The network takes in a continuous 5D coordinate representing the spatial location and viewing direction of a point in the scene and outputs the volume density and emitted radiance at that point. The algorithm uses volume rendering techniques to synthesize new views of the scene by querying the network along camera rays and projecting the output colors and densities into an image. While the results show that they outperform most previous representations, one limitation of the original NeRF approach is its long training and rendering time, which makes it difficult to use in interactive or real-time applications. There are two directions towards tackling this problem: the first is to efficiently train, store, and evaluate a single neural network that represents 3D scenes and appearances as methods described in \cite{donerf}, and the second is to use multiple deep neural networks to accelerate the rendering process, with some representatives like \cite{kilo}.

\subsection{Efficient NeRF Representation and Rendering}
The limitation of vanilla NeRF is clear: the evaluation of each sample requires querying the large underlying neural network, which can take an excessive amount of time. Additionally, the evaluation of the color of each pixel involves thousands of such queries, further increasing the computational burden. To address this issue, various methods have been developed to improve the efficiency of NeRF evaluation and reduce the amount of time required, thus allowing for faster and more efficient NeRF rendering.

There are many different approaches towards a more refined and high-performance NeRF during the progress, like \cite{donerf,supervise,baking,plen,direct,autoint,fast,mixture}, all trying to push the performance of training and inference of a single NeRF MLP to the next level. Some work like \cite{plen} uses more space to calculate and bake intermediate results in storage to trade time consumption. \cite{donerf}, on the other hand, involves using a classification network to predict sample locations foreach view ray, adaptively calculate and collect more important samples, resulting in a compact dual network to reduce inference cost without requiring additional memory for caching or acceleration structures. The AdaNeRF \cite{AdaNeRF} used adaptively placed samples ideas and pushed performance further. The \cite{mip} goes along this way to further reduce the number of samples needed to be inferred during rendering. Instead of rendering multiple rays through a single pixel from the film, it reasons about the 3D conical frustum defined by a camera pixel. By approximating the frustum with a multivariate Gaussian, they can render anti-aliased conical frustums instead of rays efficiently, avoiding most neural network inferences.  

Another group of work, like \cite{instant}, developed a new input encoding scheme that used a multiresolution hash table of trainable feature vectors optimized through stochastic gradient descent, which allows using a smaller neural network for neural graphics primitives while maintaining high-quality image synthesis. These works can be considered orthogonal to ours, as most techniques can also be applied to the following group of techniques, which is to use multiple NeRFs to represent a single scene for efficient training and rendering. 

Some work with multiscale in mind is BungeeNeRF described in \cite{bungeenerf}. The rendering process starts with a shallow base block to fit the distant views, and as the training progresses, new blocks are added to capture more details in the increasingly closer views. Our approach also incorporates multiscale ideas in the distillation and sampling process while it can also benefit from orthogonal methods that try to reduce sample counts as well. Works like DONeRF \cite{donerf} or AdaNeRF \cite{AdaNeRF} are single NeRFs with adaptive sampling, while we utilized the intrinsic adaptivity of our constructed multi-NeRF structure, which is cooperated with the multi-NeRF inference framework. This can achieve much faster results with some trade-offs in memory footprint.

Our method also employs multiple NeRFs to represent a single continuous scene, leveraging a wider range of scaling techniques. The concept of using multiple NeRFs for a single scene is not novel. KiloNeRF \cite{kilo} was first introduced as a popular multi-NeRF method that divides the scene bounding box into thousands of smaller uniform grids, each represented by a small MLP compared to the original single NeRF. Scene knowledge is distilled from the global single NeRF into each of these smaller MLPs. Evaluating smaller MLPs can be orders of magnitude faster than the inference time of the original NeRF, allowing multi-NeRF methods to achieve nearly linear performance gains. Adaptive Voronoi NeRFs \cite{adavo} employ Voronoi grids instead of axis-aligned bounding boxes to create scene subdivisions, simplifying the calculation of which MLP each sample belongs to. Works such as \cite{block} and \cite{nsvf} also utilize the multi-NeRF concept to represent larger scenes with unbalanced scene complexity. 

\section{Method}

We developed a new training and rendering framework for our method that leads to efficient NeRF rendering. The first component utilizes adaptive subdivision for our multi-NeRF structure definition and density-aware multi-NeRF construction. The second component is an effective sampling scheme that efficiently exploits parallelism through a sort-compaction algorithm, which also utilizes the hierarchical NeRF for a mip-map style approximation.

In the following sections, we will describe the multi-NeRF construction phase, training phase, and rendering phases of our framework. We will also discuss related discoveries and ablation tests in the following sections.

\subsection{Background}

The original work by \cite{nerf} describes NeRF as a scene representation using an MLP with spatial position as input and direction-dependent color and density as output. NeRF can be defined as a continuous 5D vector-valued function, denoted as $f: \mathbb{R}^3 \times \mathbb{R}^2 \rightarrow \mathbb{R}^3 \times \mathbb{R}$. The function takes a 3D location $\mathbf{x} = (x, y, z) \in \mathbb{R}^3$ and a 2D viewing direction $(\theta, \phi) \in \mathbb{R}^2$ as input arguments. The output of the function is a combination of an emitted color $\mathbf{c} = (r, g, b) \in \mathbb{R}^3$ and a volume density $\sigma \in \mathbb{R}$. In applications, the direction is represented by a three-dimensional direction vector, making NeRF a continuous function represented by an MLP (multi-layer perceptron) performing the following mapping: $F_\Theta : (\textbf{x}, \textbf{d}) \rightarrow (\textbf{c}, \sigma) \nonumber$, from the position/direction pair to a color/density pair.

In the training and rendering stages, samples are collected by ray marching from the camera, and the samples and corresponding colors are composited using traditional volumetric rendering techniques to form the color of the ray, which is then returned to the pixel plane for rendering. Our work also includes a modification of this sample accumulation process, which will be introduced in a later section.

Since samples forming the color of a ray can be decomposed without any dependencies and are fully differentiable, the network itself can also be decomposed linearly by simply making subdivisions so each will represent colors of the scene within a specific bounding area. This approach of using multiple NeRFs to represent the scene, rather than relying on a single Mega-NeRF, was introduced by \cite{kilo} and also used by \cite{block}. This decomposition using multi-NeRF instead of a single meta-NeRF allows each NeRF $F_{\Theta_i}(x, d)$ to independently capture color and directional values, enabling a more efficient scene representation with a very small NeRF, without the need for a single Mega-NeRF encompassing the entire scene.

From this observation, we introduce adaptive multi-NeRF: each subdivision can be scene-aware instead of being regularly subdivided, as in KiloNeRF \cite{kilo}. A large bounding box enclosing fewer geometries can be effectively equal to a smaller portion full of scene details, from a scene complexity perspective. We divide the scene in a BSP tree manner (KD-tree in this case), with the scene complexity of each division approximately similar to other nodes on the same layer of the tree. In this case, we expect fewer NeRFs to be created.

In this work, we employ a KD-tree for spatial partitioning, as opposed to using a uniform grid. Each KD-tree node corresponds to an axis-aligned bounding box (AABB) and a NeRF model with an independent set of parameters $\theta(i)$, where $i$ denotes the node index in the tree. We use a simple binary code to represent the node index, where each bit indicates whether the node is located to the left or right. The KD-tree-based spatial partitioning provides a more adaptive and efficient way of dividing the scene compared to a uniform grid, enabling the handling of complex scenes with varying density and structure while maintaining computational efficiency.

In the following sections, we will discuss the construction and rendering of our multi-NeRF structure and how it operates within the hierarchical structure. Although mapping samples in a hierarchical structure is more complex than using a regular grid, it can be accomplished without significant additional effort in the implementation, provided that GPU-friendly techniques are employed.

\subsection{Adaptive Subdivision of a Scene and Multi-NeRF Method}

An intuitive consideration of NeRF's representation capability is that larger MLPs are better suited for more complex scenes. We postulate that an MLP of a specific width and depth has a certain capability to represent a scene up to a limited error bound, given the same training conditions and dataset. In previous work such as \cite{instant} and \cite{donerf}, it has been observed that more complex scenes necessitate larger MLPs to capture details without compromising accuracy. Figure \ref{fig:complex} illustrates how the representation capacity of an MLP used in a NeRF is influenced by its size and the complexity of the scene. A logical approach would be to assign more complex MLPs to intricate parts of the scene and smaller NeRFs to simpler parts. However, we aim to maintain a single MLP size for all NeRFs to ensure regular GPU computations, but to change the size of each subdivision in order to fit the MLPs to a corresponding size of the scene.

\vspace{1em}

\begin{minipage}{\linewidth}
  \centering

    \begin{tabular}{l|llll}
    \textbf{RGB network size} & \textbf{32x4} & \textbf{64x8} & \textbf{128x8} & \textbf{256x8} \\ \hline
    \textbf{Material}         & 33.6          & 34.2          & 34.7           & \textbf{35.2}  \\
    \textbf{Lego}             & 34.2          & 35.7          & 36.5           & \textbf{36.4}  \\
    \textbf{Ueno}             & 30.0          & 29.8          & 31.4           & \textbf{32.1}  \\
    \textbf{Dragon}           & 27.5          & 28.9          & 29.5           & \textbf{29.9} 
    \end{tabular}
    
    \begin{figure}[H]
    \centering
    \includegraphics[width=0.8\textwidth]{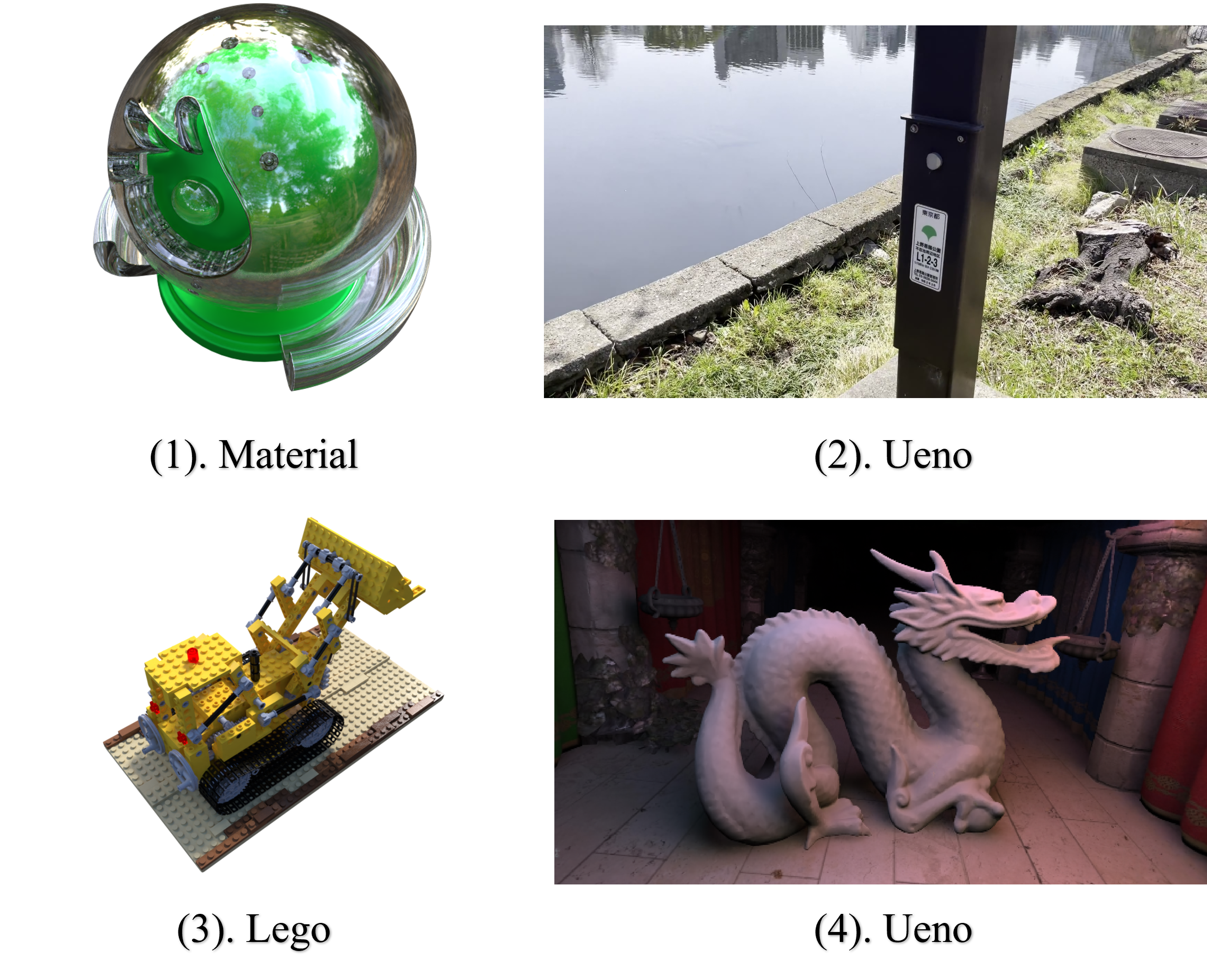}
    \caption{Network size and PSNR (10k iteration) of four different scenes represented by different size of MLPs with frequency positional encoding. }
    \label{fig:complex}
  \end{figure}
\end{minipage}
\vspace{1em}

Evaluating the accuracy of small NeRFs representing only a part of the scene cannot be achieved using the original image and their PSNR, as these NeRFs must work in conjunction to render the complete image. Consequently, we adopt a different method to estimate whether a small NeRF is sufficient for representing a scene portion. If the small NeRF lacks the necessary representation capacity, we subdivide the node, train the distilled sub-network within a predefined time limit, and repeat the process. To accomplish this, we must estimate scene complexity and identify the more intricate parts that require further subdivision.

\subsection{Scene Complexity Estimation}

We initiate our approach by performing an adaptive subdivision of the scene based on an estimated density grid. To achieve an accurate estimation, we first train a larger NeRF with a width of 256 and a depth of 8, following the methodology presented in the original NeRF paper \cite{nerf}. We then integrate this with the hash encoding technique from \cite{instant} to enhance the representational capacity of the global NeRF. In this context, we refer to this NeRF as Mega-NeRF, which denotes the largest global NeRF that is crucial for the subsequent stages of our methodology. It is important to note that our use of the term Mega-NeRF differs from that found in other works.

Based on our examination of scene complexity and neural network capacity, we recognize a correlation between a specific measurement, the input of the parameterized scene, and the current NeRF setting. We employ a density estimation heuristic to guide the construction of the next hierarchical level.

Let $\mathcal{P}$ represent a point cloud containing uniformly sampled points from the original axis-aligned bounding box (AABB) of the scene. We process this point cloud using our initial Mega-NeRF model, $N$, which maps each point $p_i$ to a corresponding volume density, $\rho_i$, and a view-dependent emitted radiance value, $L_i(\omega)$, where $\omega$ denotes the viewing direction. The density and color results are then normalized and utilized in subsequent steps.

\begin{figure*}[!tbp]
\centering
\includegraphics[width=\textwidth]{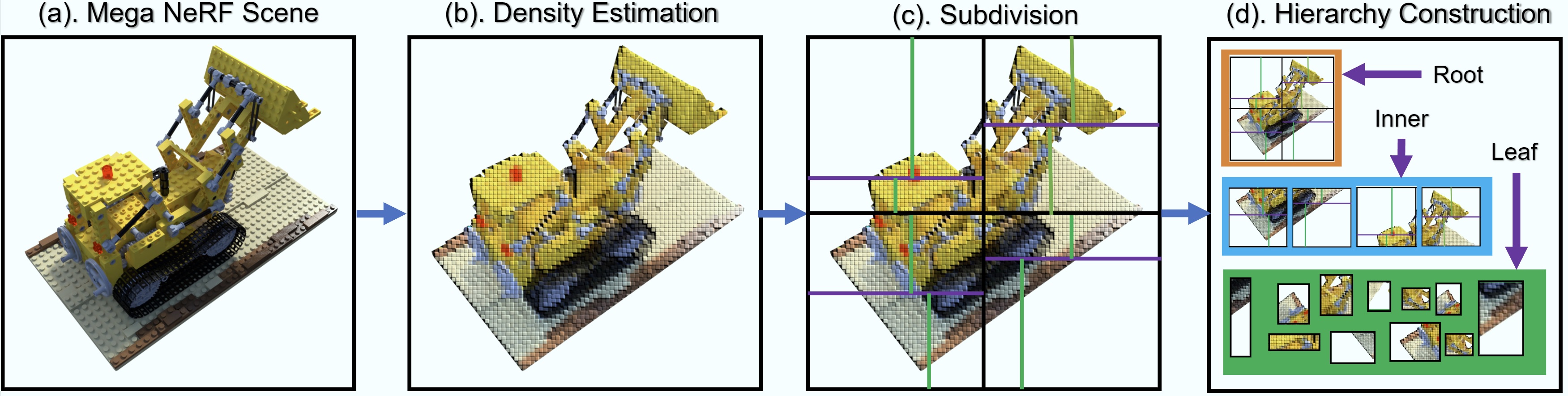}
\caption{The density-aware scene subdivision process for hierarchical multi-NeRF. (a) Mega-NeRF is trained in a conventional manner using \cite{nerf} or \cite{mip} to establish the knowledge base Mega-NeRF for the scene. (b) We subsequently use uniform samples within the axis-aligned bounding box of the scene and the estimated results obtained by querying the original Mega-NeRF. (c) During the subdivision phase, the scene is partitioned based on the estimations. Each section is represented by a significantly smaller MLP compared to the global NeRF, and (d) all inner and leaf nodes are retained, with each node serving as an AABB for a specific small NeRF.}
\label{fig:flow}
\end{figure*}

Using the density estimation heuristic, we effectively subdivide the scene to form a hierarchical structure that adapts to varying levels of complexity. In this manner, we create a series of smaller NeRF models, each representing a specific region of the scene with its corresponding AABB. This approach enables efficient scene representation and rendering, with the smaller NeRF models working in concert to generate a high-quality image while minimizing computational overhead. 

\subsection{Density-aware Adaptive Subdivision}

If we only keep the leaf node and use a regular scene subdivision, the hierarchy will be reduced to the structure described in \cite{kilo}. Based on some of the ablation tests we conducted on the same framework with both regular and adaptive subdivisions, using an adaptive tree structure (as in our implementation, a shallow KD-tree structure) almost always performs better than a regular grid.

Subsequently, we form the point cloud $\mathcal{Q}$, with each point bearing the density estimation obtained from the previous step. With the normalized point cloud $\mathcal{Q}$, we proceed to construct a KD-tree, denoted by $T$, to efficiently organize the data. The KD-tree construction process follows a recursive procedure, where at each level, we split the data based on the dimension that yields the median of the cumulative density values. This ensures a balanced partitioning of the points in $\mathcal{Q}$. An overview of the process is shown in Figure \ref{fig:flow}. Here is also a simple description of the subdivision and hierarchy construction process:

\begin{verbatim}
Define BuildKDTree (input: point cloud, dimensions)
1.   If TestLeaf(point cloud) passed, return leaf;
2.   Sort points by coordinates in chosen dimension;
3.   Compute median, Create new internal node; 
4.   Partition point cloud into two subsets;
5.   Recursively build left and right subtrees;
\end{verbatim}

A crucial aspect of our methodology is the TestLeaf() procedure, which determines when to stop further subdivision. Several factors must be considered, as they can significantly impact performance during subsequent stages when managing layers. Increasing the number of layers implies assigning MLPs to smaller portions of the scene. While this has the potential to enhance accuracy by using MLPs with the same dimensions to represent smaller scenes and effectively increasing capacity, there are two apparent disadvantages to creating a larger number of MLPs: (1) Storage requirements grow, as a balanced KD-tree can have an order of $O(2^{d+1} - 1)$ with $d$ as its depth, resulting in substantial storage consumption. For GPU-based applications, this can quickly become unacceptable. (2) Overly fine-grained subdivisions can lead to a significant reduction in parallelism. As data in each finely partitioned area becomes sparse, parallelism and occupancy decrease. Consequently, it is essential to strike a balance between subdividing the problem into manageable subproblems and maintaining an appropriate level of parallelism to optimize the utilization of available computational resources.

\begin{figure}[H]
\centering
\includegraphics[width=0.5\textwidth]{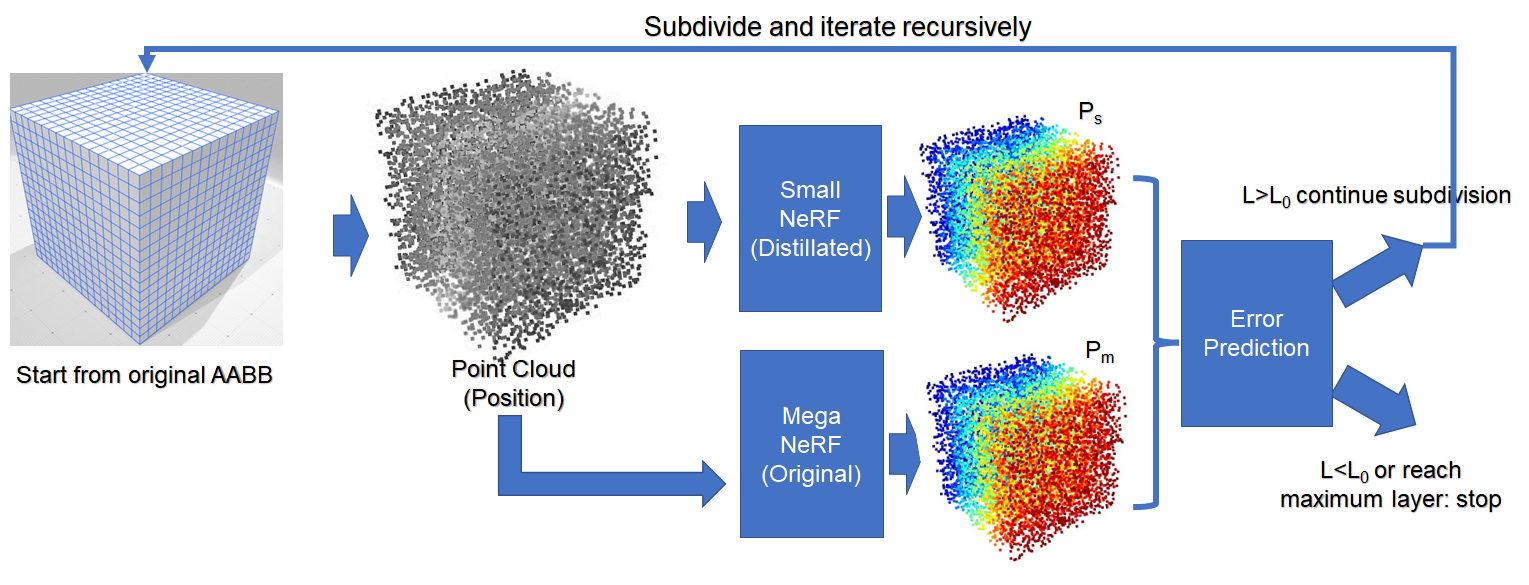}
\caption{The stopping criterion for KD-tree subdivision. The construction process terminates and returns a leaf node when specific conditions are met. The criterion compares the average PSNR obtained after a predefined number of distillation iterations to the original tree and stops further subdivision when the difference falls below a predetermined threshold. This method ensures an optimal balance between model accuracy and computational efficiency.}
\label{fig:kd_tree_stopping_criterion}
\end{figure}

In this section, we introduce the criterion for determining the stopping point in the KD-tree subdivision process. The proposed criterion involves constructing a NeRF model corresponding to each node in the KD-tree. These NeRF models, with predefined dimensions such as a width of 32 and a depth of 4, represent the scenes enclosed by the axis-aligned bounding boxes (AABBs) associated with the respective KD-tree nodes.

Throughout the subdivision process, each NeRF model undergoes a distillation phase, learning parameters based on the Mega-NeRF, which represents the entire global scene encompassed by the KD-tree or, equivalently, corresponds to the root node's associated MLP with an original size of a width of 256 and a depth of 8. The distillation process is executed for a maximum of $N_{max}$ iterations, after which the average PSNR (Peak Signal-to-Noise Ratio) is compared to the value obtained from the original tree. If the difference in PSNR falls below a predetermined threshold, such as 1\%, the subdivision process should be halted for the current node. This decision is based on the observation that the MLP size and scene density are well-matched, and further subdivision would not yield any significant improvements in representation.

By integrating the distillation and subdivision processes, our proposed method allows for efficient hierarchical execution of samples while simultaneously adapting the model complexity to the local scene density. This approach ensures a balance between model accuracy and computational efficiency, making it well-suited for large-scale scene reconstruction tasks.

\subsection{Adaptive Hierarchical Sampling and Parallel Scheduling for Rendering}

In this part, we describe how to use adaptive multi-NeRF method for rendering. After the construction and distillation of the hierarchy, we have a multi-NeRF structure with each inner node and leaf node bearing a part of scene representation knowledge within its MLPs. During rendering, rays will be intersecting the BSP tree we constructed, and samples will be generated, inferred an composited to form the color of each ray and the pixel it represents. We choose to record intervals that each ray intersects with the KD-tree, and the fill samples in the intervals in a uniform way. A psudo-code overview of traversing rays, generate intervals and fill the samples is shown in Algorithm 1.

\begin{algorithm}
\caption{KD-tree traversal and sample generation}
\begin{algorithmic}[1]
\For{each ray in raylist with index $i$ in parallel}
    \State \Call{Traverse}{$\text{KDtreeNode}$, $\text{ray}$, $\text{current\_t}$}
\EndFor
\Procedure{Traverse}{$\text{KDtreeNode}$, $\text{ray}$, $\text{current\_t}$}
    \If{$(\text{KDtreeNode is inner node \& HCheck(KDtreeNode)})$}
        \State \Call{Traverse}{$\text{KDtreeNode.left}$, $\text{ray}$, $\text{current\_t}$}
        \State \Call{Traverse}{$\text{KDtreeNode.right}$, $\text{ray}$, $\text{current\_t}$}
    \EndIf
    \If{$(\text{intersect(KDtreeNode, ray, current\_t})$}
        \State Get intersection points of leaf\_node AABB ($\text{current\_t}$, $t_1$)
        \State $\text{intervallist[rayID]} \gets (\text{current\_t}, t_1)$
        \State $\text{current\_t[rayID]} \gets t_1$
        \State \textbf{return}
    \Else
        \State Mark rays as inactive in parallel
        \State \textbf{return}
    \EndIf
\EndProcedure
\For{each interval generated by ray $i$ in parallel}
    \State \Call{Sampling}{$\text{intervals}[i]$}
\EndFor
\Procedure{Sampling}{$\text{intervals}[i]$}
    \State $\text{interval} \gets \text{normalize(intervals[i])}$
    \State $n \gets \text{calculate\_sample\_size(interval[i])}$
    \For{$j = 1$ to $n$}
        \State $\text{sample} \gets \text{generate\_low-discrepancy\_sample()}$
        \State $\text{samplelist}[i * \text{max\_cell\_count} + j] \gets (\text{sample}, \text{rayID})$
    \EndFor
\EndProcedure
\end{algorithmic}
\end{algorithm}

Samples are generated by marching the rays and recorded by the rayID and node index. Uniform random samples or low discrepancy samples are filled within each intersection interval. Here, we explain the HCheck procedure, which determines whether interval outputs are made on the inner node or the leaf node. Obtaining intervals solely for the leaf node of the hierarchy may reduce performance, as distant dense nodes may also be sampled intensively. As the distance between the camera and the scene increases, the number of query points in a specific grid decreases due to the expanding separation between adjacent ray sample points. Figure \ref{fig:teaser_image} provides an illustration of this concept.

The current execution scheme could lead to a significant workload problem based on the data collected during the process. Launching kernels with very few samples within the AABB may cause hardware usage to drop significantly. One solution is to use a mega CUDA kernel to organize parallel scheduling, as described in the KiloNeRF implementation \cite{kilo}. However, such a strategy for increasing parallelism is not well-suited for our task, which directly utilizes the fast small MLP library. Although this scheduling reduces kernel call overheads, thread divergence during execution can still cause occupancy to drop.

\begin{figure}[h]
\includegraphics[width=0.5\textwidth]{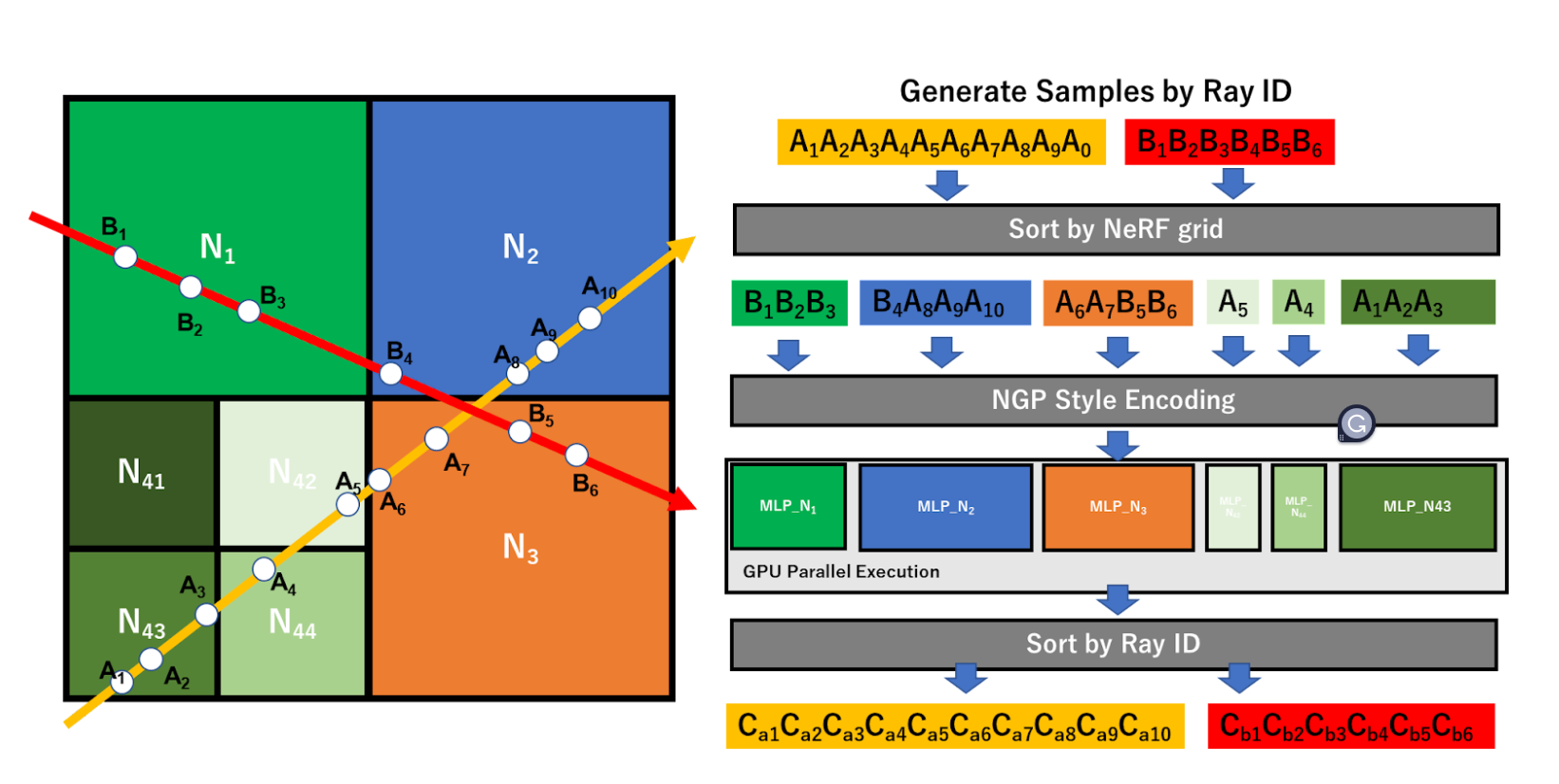}
\centering
\caption{Samples are generated through the ray-KDtree intersection test as described in the sampling section. Parallel groups, which are executed in parallel, are organized in a spatial grid containing the samples.}
\label{fig:parallel}
\end{figure}

Samples, along with their payload, rayID, and nodeID, are then sorted and inferred in the corresponding neural network. This process must be carefully designed to minimize the memory footprint and the number of kernel calls on the GPU. Sorting is a comparatively less expensive operation for GPU parallel execution, so we choose it for scheduling and maximizing parallelism. Figure \ref{fig:parallel} shows the scheduling and inference process for all samples generated during the sampling phase.

\begin{figure}[h]
\centering
\includegraphics[width=0.5\textwidth]{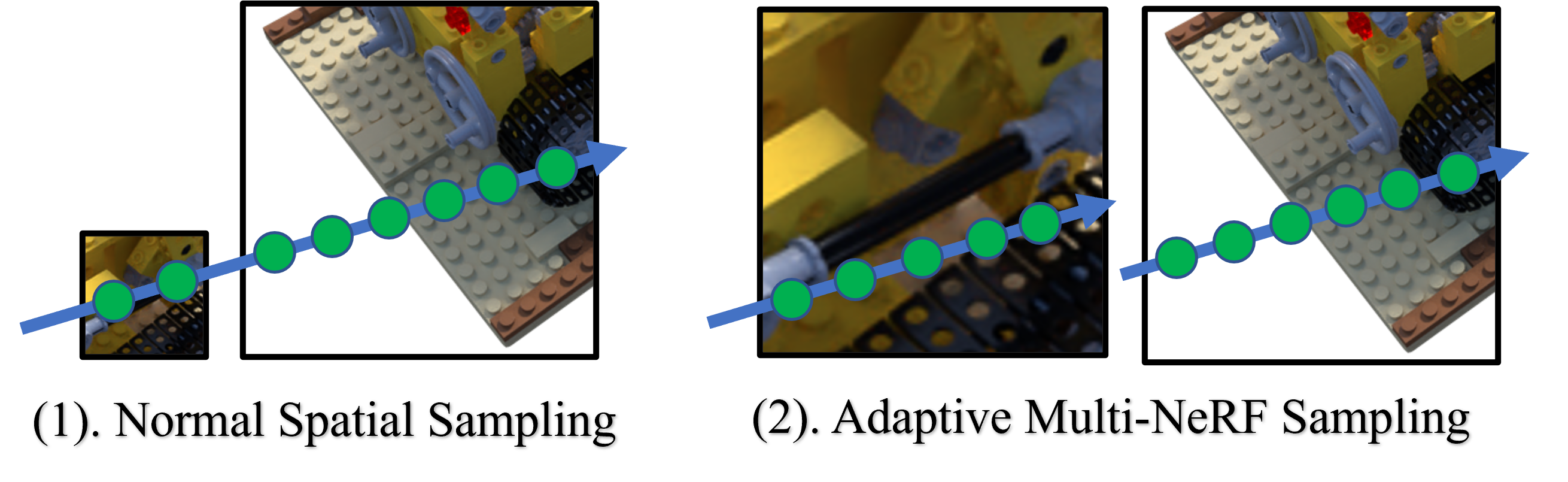}
\caption{(1) Without adaptive methods applied, samples are generated proportionally to the distance of intervals along the ray, leading to a small number of samples in complex regions with small spatial extensions. (2) In our adaptive multi-NeRF structure, all intervals are normalized to a unit space, and samples are generated using an equal distribution strategy. This ensures that regions with similar complexity receive a comparable number of samples, regardless of their spatial dimensions.}
\label{fig:prop}
\end{figure}

Here, we can also confirm the intrinsic adaptivity of utilizing adaptive multi-NeRF: the intervals of large nodes will be significant in spatial terms, but they will be treated the same as intervals in a smaller size node. This is because they essentially represent a part of the scene with similar complexity, requiring similar sample counts for capturing and integrating in the volumetric color compositing. See Figure \ref{fig:prop} for reference.

Recursion will not be very deep, as we currently limit the multi-NeRF depth to 10. When the size of multi-NeRF increases, the overhead of parallel scheduling may not be sufficient to compensate for the gain, and samples in each cell may not be large enough to saturate GPU execution.

\section{Evaluation}
\subsection{Experiments Setup}
We implemented our method within a modified framework of Instant-NGP, as proposed by \cite{instant}. For most ablation tests and comparison we employed hash encoding of \cite{instant} to reap the benefits of rapid convergence. The network architecture of our adaptive multi-NeRF is similar to the small NeRF described in \cite{kilo} but to accommodate the hash encoding, we modified the layout to include concatenated density and RGB networks. We used the same dimensions for both networks in our tests, representing them by their width and inner layer depth. For the tree structure of the scene subdivision we chose a maximum layer of 10, ensuring that the scene would not be subdivided much more beyond 1k. This design choice differs from KiloNeRF \cite{kilo}, as we continue to employ the finer density grid in \cite{instant} for hash-grid encoding, early ray termination, and empty space detection while KiloNeRF utilized MAGMA framework for better parallelization performance.

We compared the performance based on the 20k iteration results from the Mega-NeRF training with an MLP of 256x8 similar to the vanilla NeRF in \cite{nerf}. Our training configuration followed that of Instant-NGP in \cite{instant}, except when training the overworld/global Mega-NeRF. In this case, we used NeRF-PyTorch to train an accurate frequency encoding version and then applied distillation to train an RGB network of 128x6 as a comparison to the global network of Instant-NGP. Regarding the KiloNeRF implementation we compared the results with our implementation only that to use a regular subdivision to divide scenes into regular grids, instead of using the BSP tree.

We employed several original NeRF datasets, encompassing both synthetic and real-world scenes, for smaller scene tests. Additionally, we utilized two scenes from \cite{Pilkington2022} for larger and more imbalanced scene tests. We created the Ueno and Dragon scene and also modified the environment map and refraction parameters of the Material scene.

\subsection{Ablation studies}
We did some necessary ablation study to evaluate that some of our assumptions are valid. Table \ref{tab:Ablation} is a simple summary.

\begin{table}[h]
\centering
\resizebox{0.5\textwidth}{!}{%
\begin{tabular}{|l|l|l|l|}
\hline
                                     & Avg. Frame Time (ms)↓ & Avg. PSNR↑    & Avg. Samples   per ray \\ \hline
Ours   without irregular subdivision & 52.6                  & 26.3          & 77.1                   \\ \hline
Ours without hierarchical   sampling & 77.8                  & 26.3          & 88                     \\ \hline
Complete setup                       & \textbf{45.7}         & \textbf{26.7} & \textbf{56}            \\ \hline
\end{tabular}%
}
\caption{Ablation study}
\label{tab:Ablation}
\end{table}
We investigated the effectiveness of irregular subdivision using KD-trees, as in our method, in comparison to regular subdivision, where scenes are evenly divided into grids and each bounding box is represented by a small NeRF as in KiloNeRF \cite{kilo}. Notably, we utilized far fewer NeRFs in both cases. Our tests revealed that irregular scene subdivision resulted in approximately 32\% fewer samples generated per ray, slightly improved image quality, and overall enhanced performance as shown in the frame generation time.

The implementation of adaptive hierarchical sampling further reduced the sample count by around 42\%. Since we adopted the Instant-NGP framework from \cite{instant} to realize our concept, kernel calls were minimized as the networks handling scattered samples were reduced. Samples farther from the ray were more likely to be evaluated by an inner node instead of leaf nodes, leading to a more efficient network inference process.

\subsection{Comparisons}
In this section, we analyze the performance of our method in comparison to other state-of-the-art techniques. Table \ref{tab:com} presents a comprehensive summary of the results, including memory usage, frame time, PSNR, and average samples per ray for various scenes such as Material, Ueno, Lego, Dragon, House1, and House2.

Our method demonstrates competitive frame times, ranging from 16.5 ms to 91 ms across the different scenes for generating native 500x500 pixel images on a single NVIDIA RTX 8000. On average, our method is about 30 percent faster than approaches that do not apply our adaptive subdivision multi-NeRF techniques. It is worth noting that the Material scene is replete with highly view-dependent reflections, resulting in consistently high density throughout the volume rendering process. In this case, the adaptive subdivision reverts to a BSP tree-style regular grid subdivision, and a single Mega-NeRF proves to be a more suitable solution.

When Adaptive Hierarchical Sampling (AHS) is not employed in our method (Ours w/o AHS), the frame time increases, ranging from 45.6 ms to 112 ms. This observation underscores the effectiveness of AHS in reducing frame generation times.

The table primarily showed that the proposed method provides significant improvements in frame generation times while maintaining competitive PSNR values compared to state-of-the-art techniques. By adaptively subdividing the scene and employing hierarchical sampling, our approach ensures a balance between model accuracy and computational efficiency, making it a promising solution for large-scale scene reconstruction tasks. However, we also noticed how storing every inner nodes for a mip-map style sampling and inference can be very costly in memory usage, which will be conpensated by using only leaf nodes and avoid hierarchical sampling as a whole.

\begin{table*}[t]
\resizebox{\textwidth}{!}{%
\begin{tabular}{lllllllll}
\hline
512x512   no dynamic res & Memory {[}MB{]} & \multicolumn{1}{l|}{}                     & Material                           & Ueno                               & Lego                               & Dragon                             & House1                             & House2        \\ \hline
Ours                     & 495             & \multicolumn{1}{l|}{Frame Time (ms) ↓}    & \multicolumn{1}{l|}{32}            & \multicolumn{1}{l|}{\textbf{16}}   & \multicolumn{1}{l|}{\textbf{29}}   & \multicolumn{1}{l|}{\textbf{31}}   & \multicolumn{1}{l|}{\textbf{74}}   & \textbf{91}   \\ \cline{1-3}
                         &                 & \multicolumn{1}{l|}{PSNR ↑}               & \multicolumn{1}{l|}{27.1}          & \multicolumn{1}{l|}{24.8}          & \multicolumn{1}{l|}{35.0}          & \multicolumn{1}{l|}{24.9}          & \multicolumn{1}{l|}{22.5}          & 24.1          \\ \cline{3-3}
                         &                 & \multicolumn{1}{l|}{Avg. Samples per ray} & \multicolumn{1}{l|}{24}            & \multicolumn{1}{l|}{57}            & \multicolumn{1}{l|}{39}            & \multicolumn{1}{l|}{31}            & \multicolumn{1}{l|}{90}            & 97            \\ \cline{1-3}
Ours w/o AHS             & 495             & \multicolumn{1}{l|}{Frame Time (ms) ↓}    & \multicolumn{1}{l|}{45}            & \multicolumn{1}{l|}{57}            & \multicolumn{1}{l|}{55}            & \multicolumn{1}{l|}{68}            & \multicolumn{1}{l|}{128}           & 112           \\ \cline{1-3}
                         &                 & \multicolumn{1}{l|}{PSNR ↑}               & \multicolumn{1}{l|}{27.0}          & \multicolumn{1}{l|}{24.9}          & \multicolumn{1}{l|}{35.0}          & \multicolumn{1}{l|}{24.4}          & \multicolumn{1}{l|}{21.6}          & 25.2          \\ \cline{3-3}
                         &                 & \multicolumn{1}{l|}{Avg. Samples per ray} & \multicolumn{1}{l|}{55}            & \multicolumn{1}{l|}{69}            & \multicolumn{1}{l|}{70}            & \multicolumn{1}{l|}{72}            & \multicolumn{1}{l|}{120}           & 145           \\ \cline{1-3}
KiloNeRF*                & 225             & \multicolumn{1}{l|}{Frame Time (ms) ↓}    & \multicolumn{1}{l|}{39}            & \multicolumn{1}{l|}{50}            & \multicolumn{1}{l|}{52}            & \multicolumn{1}{l|}{71}            & \multicolumn{1}{l|}{94}            & 107           \\ \cline{1-3}
                         &                 & \multicolumn{1}{l|}{PSNR ↑}               & \multicolumn{1}{l|}{26.2}          & \multicolumn{1}{l|}{22.8}          & \multicolumn{1}{l|}{33.1}          & \multicolumn{1}{l|}{23.6}          & \multicolumn{1}{l|}{22.0}          & 24.1          \\ \cline{3-3}
                         &                 & \multicolumn{1}{l|}{Avg. Samples per ray} & \multicolumn{1}{l|}{75}            & \multicolumn{1}{l|}{82}            & \multicolumn{1}{l|}{42}            & \multicolumn{1}{l|}{64}            & \multicolumn{1}{l|}{80}            & 120           \\ \cline{1-3}
Instant-NGP              & 72**            & \multicolumn{1}{l|}{Frame Time (ms) ↓}    & \multicolumn{1}{l|}{\textbf{29}}   & \multicolumn{1}{l|}{26}            & \multicolumn{1}{l|}{46}            & \multicolumn{1}{l|}{49}            & \multicolumn{1}{l|}{95}            & 111           \\ \cline{1-3}
                         &                 & \multicolumn{1}{l|}{PSNR ↑}               & \multicolumn{1}{l|}{26.1}          & \multicolumn{1}{l|}{25.8}          & \multicolumn{1}{l|}{33.1}          & \multicolumn{1}{l|}{24.2}          & \multicolumn{1}{l|}{22.3}          & 24.0          \\ \cline{3-3}
                         &                 & \multicolumn{1}{l|}{Avg. Samples per ray} & \multicolumn{1}{l|}{65}            & \multicolumn{1}{l|}{117}           & \multicolumn{1}{l|}{52}            & \multicolumn{1}{l|}{66}            & \multicolumn{1}{l|}{76}            & 111           \\ \cline{1-3}
NeRF                     & \textbf{5}      & \multicolumn{1}{l|}{Frame Time (ms) ↓}    & \multicolumn{1}{l|}{$\sim$seconds} & \multicolumn{1}{l|}{$\sim$seconds} & \multicolumn{1}{l|}{3750}          & \multicolumn{1}{l|}{$\sim$seconds} & \multicolumn{1}{l|}{$\sim$seconds} & $\sim$seconds \\ \cline{1-3}
                         &                 & \multicolumn{1}{l|}{PSNR ↑}               & \multicolumn{1}{l|}{\textbf{29.9}} & \multicolumn{1}{l|}{\textbf{25.4}} & \multicolumn{1}{l|}{\textbf{35.9}} & \multicolumn{1}{l|}{\textbf{-}}    & \multicolumn{1}{l|}{\textbf{23.5}} & \textbf{26.7} \\ \cline{3-3}
                         &                 & \multicolumn{1}{l|}{Avg. Samples per ray} & \multicolumn{1}{l|}{128}           & \multicolumn{1}{l|}{128}           & \multicolumn{1}{l|}{128}           & \multicolumn{1}{l|}{128}           & \multicolumn{1}{l|}{128}           & 128           \\ \hline
\multicolumn{9}{l}{* The regular grid and distillation   implemented in the NGP framework}                                                                                                                                                                                                      \\
\multicolumn{4}{l}{**   Footprint will be larger according to the step configured in ray march kernel}                      &                                    &                                    &                                    &                                    &               \\ \hline
\end{tabular}%
}
\caption{Comparison of Material, Ueno, Lego, Dragon, House1 and House2 scenes.}
\label{tab:my-table}
\end{table*}

\begin{figure*}[htbp]
  \centering
  \includegraphics[width=\textwidth]{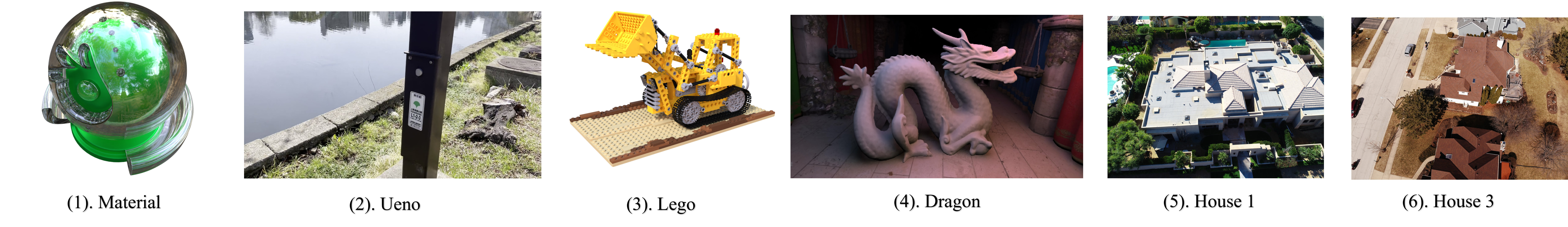}
  \caption{Test scenes used in the evaluation.}
  \label{fig:fullpage}
\end{figure*}

\subsection{Smaller NeRFs enable Efficient GPU MLP Implementations}

According to the implementation of tiny-cuda-nn provided by \cite{tiny-cuda-nn}, MLP with smaller width and depth optimized for cacheline memory access by utilizing onchip memory will significantly increase performance. Using out adaptive multi-NeRF subdivision and parallel scheduling described in previous section, we can enable the distillation of large NeRF models into smaller MLPs representing only a part of the scene, thus enabling the utilization of more efficient smaller NeRF implementation such as tiny-cuda-nn introduced by \cite{tiny-cuda-nn}. We did a test in the NeRF-pytorch framework to convert a 256x8 mega-NeRF implementation with the pytorch interface provided by tiny-cuda-nn by using our KD-tree scene subdivision without changing the sampling scheme, which means samples are still generated by the ray marching, inference and composition style. Scenes can be represented in around 128 small NeRFs of the dimension 32x4. We observed significant performance increase in the rendering wihtout much loss in the image quality as shown in table \ref{nerf_pytorch}

\begin{table}[h]
\centering
\resizebox{0.5\textwidth}{!}{%
\begin{tabular}{lll}
                   & Avg. Frame Time (ms)↓ & Avg. PSNR↑ \\
NeRF   PyTorch     & 1269                  & 32.1       \\
Multi-NeRF PyTorch & 115                   & 31.7      
\end{tabular}%
}
\caption{Use adaptive Multi-NeRF to reorganize previous Mega-NeRF methods}
\label{tab:nerf_pytorch}
\end{table}

\section{Conclusions and Future Work}

In this paper, we presented an adaptive NeRF method designed to significantly accelerate the neural rendering process for large scenes with unbalanced workloads. Our approach adaptively subdivides the scene into axis-aligned bounding boxes of varying sizes and assigns a smaller MLP to each grid as the underlying neural representation for specific parts of the scene. By optimizing the subdivision process using a guidance scene density, we ensure balanced representation capabilities for each MLP. Furthermore, we developed novel sampling and evaluation parallel scheduling techniques to maximize GPU execution capabilities by exploiting parallelism and batchifying data for efficient kernel call launches. Our method accelerates rendering from state-of-the-art by 30-40 percent without sacrificing image quality, while maintaining an acceptable memory footprint and implementation trade-offs.

Future work in this area will focus on the following aspects:

\subsection{Making the Model More Dynamic}
The proposed model can be enhanced by incorporating mip-map NeRFs to store information for each particular part, enabling a more dynamic representation. When viewed from different angles, novel views can be synthesized using NeRFs with varying levels of detail, leading to improved performance. Additionally, this approach can be combined with other adaptivity techniques to further enhance the model's performance.

\subsection{Scaling to Larger Models}
As suggested by \cite{block}, there is potential to scale this technique to represent much larger scenes with different model sizes. We believe that the adaptive multi-NeRF method can be more effective in scaling scene sizes, as it employs a scene-aware adaptive assignment method that optimizes storage and regulates the representation power of each subdivision.

\begin{acks}
We acknowledge that the arXiv version of the paper is not yet in its final form. We plan to add more evaluations and continue refining the paper in subsequent stages before submitting this work for peer review. The authors would like to express their gratitude to all anonymous reviewers who provided valuable comments and helpful suggestions, that have contributed to the improvement of this work.

The authors would like to thank members of Cygames Research for providing invaluable support for this work.
\end{acks}

\bibliographystyle{ACM-Reference-Format}
\bibliography{main}
\end{document}